\documentclass[11pt]{article}

\usepackage[final]{acl}

\usepackage{times}
\usepackage{latexsym}
\usepackage{adjustbox}
\usepackage{enumitem}
\usepackage{booktabs}
\usepackage[T1]{fontenc}
\usepackage{natbib}
\usepackage[utf8]{inputenc}

\usepackage{microtype}
\usepackage{multicol}
\usepackage{multirow}
\usepackage{multirow}
\usepackage{makecell}
\usepackage{array}
\usepackage{booktabs}
\usepackage{algorithm}
\usepackage{algpseudocode}
\usepackage{amsmath}
\usepackage{colortbl}
\usepackage{hhline}
\usepackage{inconsolata}
\usepackage{marginnote}
\usepackage{graphicx}
\usepackage{amssymb}
\usepackage{pifont}
\usepackage[flushmargin]{footmisc} 
\title{OptiVerse: A Comprehensive Benchmark towards \\ Optimization Problem Solving}

\author{Xinyu Zhang$^{1,2}$\thanks{These authors contributed equally to this work.},\; Boxuan Zhang$^{1,2\ast}$,\; Yuchen Wan$^{1,2}$,\; {Lingling Zhang}$^{1,2}$ \thanks{Corresponding author}, \;  \\
\textbf{Yixing Yao}$^{1,2}$,\; \textbf{Bifan Wei}$^{1,2}$,\; \textbf{Yaqiang Wu}$^{4}$,\;
\textbf{Jun Liu}$^{1,3}$\\
{$^{1}$School of Computer Science and Technology, Xi’an Jiaotong University}\; \\
{$^{2}$Ministry of Education Key Laboratory of Intelligent Networks and Network Security, China} \; \\
{$^{3}$Shaanxi Province Key Laboratory of Big Data Knowledge Engineering, China} \; \\
{$^{4}$Lenovo Research} \; \\
\texttt{{zhang1393869716}@stu.xjtu.edu.cn, \{zhanglling,liukeen\}@xjtu.edu.cn}
}
\begin{document}
\maketitle

\begin{abstract}
While Large Language Models (LLMs) demonstrate remarkable reasoning, complex optimization tasks remain challenging, requiring domain knowledge and robust implementation.
However, existing benchmarks focus narrowly on Mathematical Programming and Combinatorial Optimization, hindering comprehensive evaluation.
To address this, we introduce \textbf{OptiVerse}, a comprehensive benchmark of 1,000 curated problems spanning neglected domains, including Stochastic Optimization, Dynamic Optimization, Game Optimization, and Optimal Control, across three difficulty levels: Easy, Medium, and Hard.
The experiments with 22 LLMs of different sizes reveal sharp performance degradation on Hard problems, where even advanced models like GPT-5.2 and Gemini-3 struggle to exceed 27\% accuracy.
Through error analysis, we identify that modeling \& logic errors remain the primary bottleneck. 
Consequently, we propose a \textbf{Dual-View Auditor Agent} that improves the accuracy of the LLM modeling process without introducing significant time overhead.
OptiVerse will serve as a foundational platform for advancing LLMs in solving complex optimization challenges.
\end{abstract}
\section{Introduction}
Large Language Models (LLMs) have revolutionized diverse domains, such as open-ended dialogue~\cite{liu-etal-2025-persona}, mathematical reasoning~\cite{jansen2025codescientist, yan2025survey}, code generation~\cite{jansen2025codescientist}, visual understanding~\cite{zhangcofft, zhang2023rpmg, zhang2025cognitive}, temporal analysis~\cite{zhang2022tn}, and science  reasoning~\cite{zhang2025diagram, zhang2025memory}.
However, a pivotal question remains regarding the practical utility of these advancements: \textit{To what extent can LLMs translate their reasoning and programming capabilities into real-world application scenarios?}
\par
\begin{figure}[t]
\centering
\includegraphics[width=0.48\textwidth]{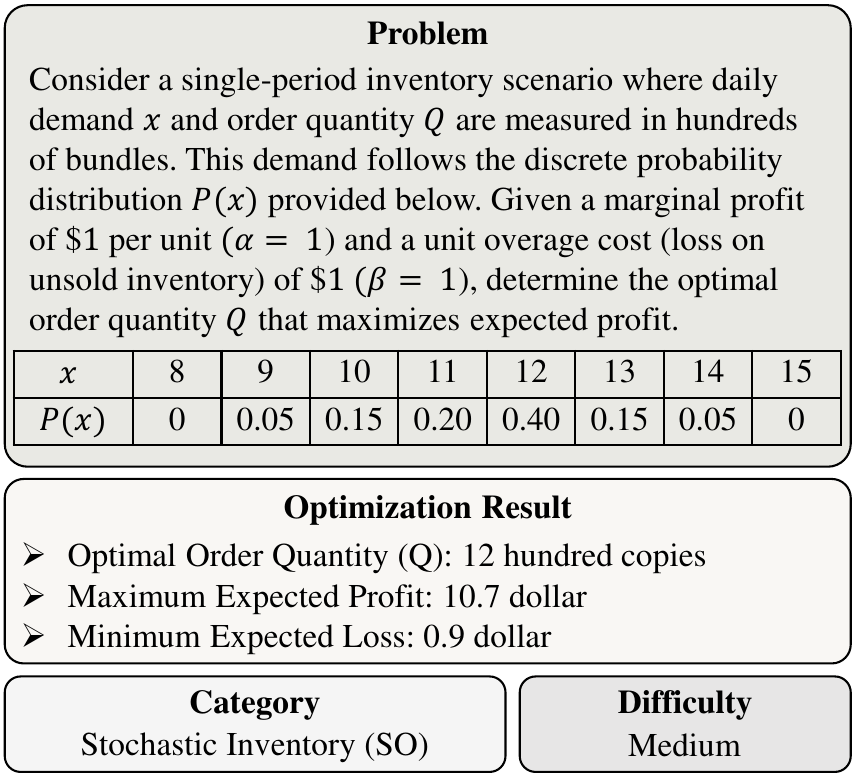}
\vspace{-15pt}
\caption{An illustrative example from \textbf{OptiVerse}.}
\label{fig:example}
\vspace{-15pt}
\end{figure}

Optimization, serving as a cornerstone discipline underpinning operations research, engineering design, and strategic decision-making~\cite{huang2025orlm}, presents a quintessential testbed for this inquiry.
Far beyond basic arithmetic, optimization demands a sophisticated synthesis of skills. 
It requires interpreting domain semantics, formulating rigorous mathematical models, and implementing executable code with solvers, as exemplified in Figure~\ref{fig:example}.
The challenge of meeting these rigorous demands, despite the general capabilities of LLMs, has ignited a growing research interest in LLM-based optimization modeling~\cite{liuoptitree}.
\par
\par
\begin{figure*}[t]
\centering
\includegraphics[width=0.98\textwidth]{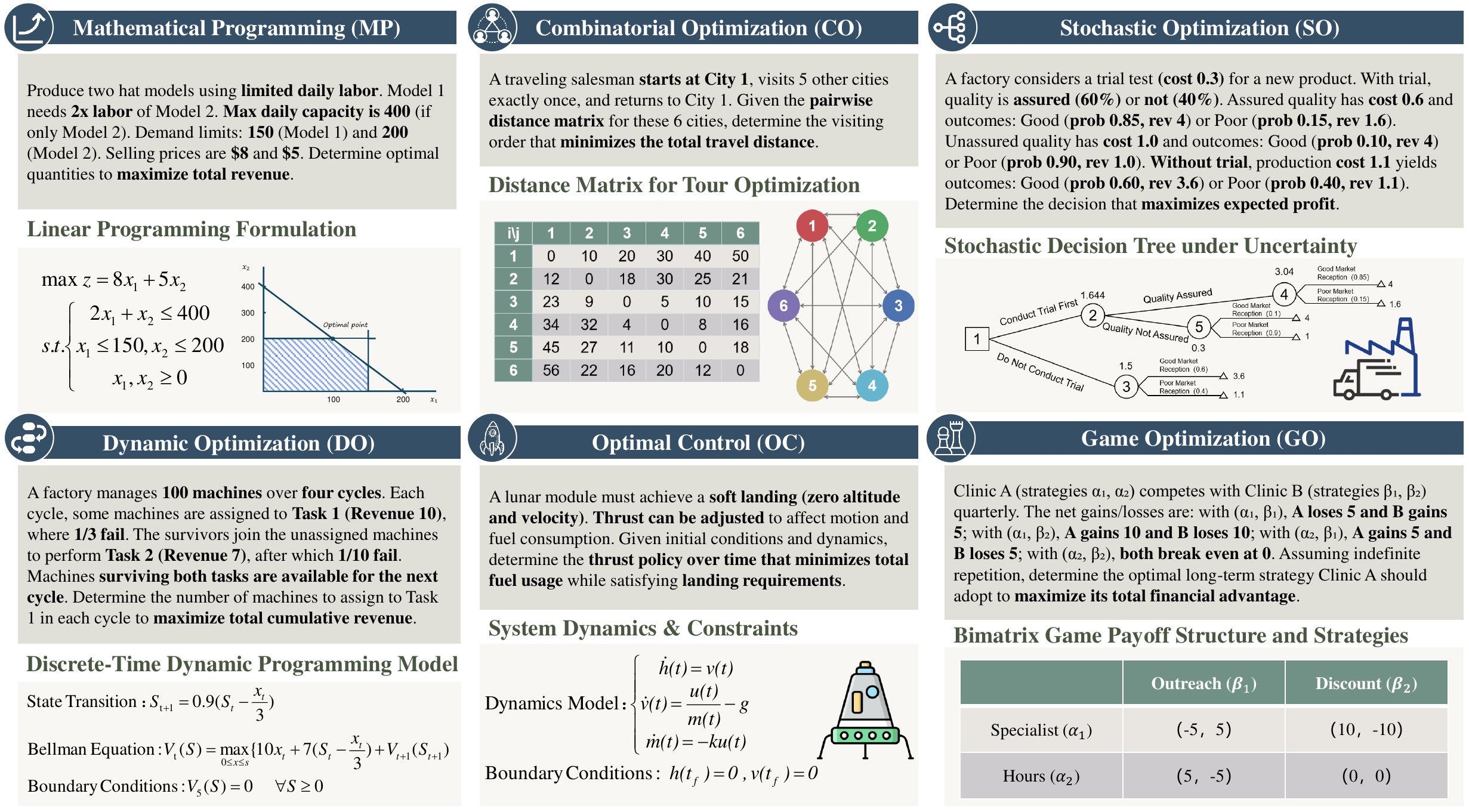}
\vspace{-5pt}
\caption{The hierarchical taxonomy of OptiVerse benchmark, which provides comprehensive coverage across six distinct optimization domains to comprehensively evaluate the diverse reasoning capabilities of LLMs.}
\label{fig:benchmark}
\vspace{-11pt}
\end{figure*}
\par
However, existing evaluation benchmarks suffer from a \textbf{distinctly narrow disciplinary scope}, which significantly hinders a comprehensive assessment of model capabilities.
Current benchmarks, such as OptiMath~\cite{luoptmath} and OptiBench~\cite{yangoptibench}, focus primarily on Mathematical Programming (MP) and Combinatorial Optimization (CO).
Crucially, they systematically neglect other essential domains ubiquitous in optimization problems.
These overlooked areas include Stochastic Optimization (SO), Dynamic Optimization (DO), Optimal Control (OC), and Game Optimization (GO) \cite{halperin2022reinforcement}.
This omission fails to reflect the true breadth of optimization capabilities, thereby limiting our analysis of LLM generalization across diverse domains.
\par
To bridge this, we introduce \textbf{OptiVerse}, a comprehensive benchmark meticulously designed to provide a rigorous, multi-dimensional evaluation of LLM-based optimization solving. 
\textbf{OptiVerse} encompasses 1,000 carefully curated problems spanning the full spectrum of six distinct optimization disciplines: MP, CO, SO, DO, OC, and GO, as show in Figure \ref{fig:benchmark}. 
This holistic coverage enables a systematic assessment of domain-specific strengths and weaknesses that single-discipline benchmarks cannot capture. 
Furthermore, to simulate the complexity of realistic applications, we organize problems into a hierarchical complexity stratification ({Easy}, {Medium}, and {Hard}), comprising 300, 400, and 300 problems, respectively.
This design enables a fine-grained analysis of how model performance degrades as problem sophistication scales.
\par
We conduct extensive and rigorous experiments to evaluate 22 Large Language Models (LLMs; evaluated on textual inputs only) across varying scales, ranging from 8B parameter models such as Qwen3-8B \cite{yang2025qwen3} to current flagship frontiers such as Gemini-3-Pro \cite{google2025gemini3} and GPT-5.2 \cite{openai2025gpt52}.
Our experiments yield critical insights:
(1) significant performance disparities exist across domains, with success rates in understudied categories (e.g., Optimal Control) often trailing common MP and CO tasks;
and (2) while current LLMs exhibit robust performance on {Easy} tasks, they struggle profoundly with {Hard} problems, where even top-tier LLMs like Gemini-3-Pro and GPT-5.2 can only reach a maximum of 27\% and 25.33\% accuracy, respectively.
\par
To investigate the reasons for these failures, we perform a systematic error analysis, revealing that \textbf{Modeling \& Logic} errors constitute the predominant bottleneck, often manifesting as silent semantic discrepancies where the modeling logic and executable code deviate from the problem intent despite successful code execution. 
To address this, we propose the \textbf{Dual-View Auditor Agent (DVA-Agent)}, which detects and repairs these subtle semantic alignment defects. 
Experimental results demonstrate that DVA-Agent significantly enhances solving capabilities; for instance, for Qwen3-235B-Instruct, it boosts success rates on {Hard} and {Medium} problems by 7.66\% and 10.5\% respectively. 
Crucially, the repair mechanism is triggered in only 32.3\% of instances, thereby maintaining computational efficiency by avoiding unnecessary iterations on already correct solutions.

\begin{table*}
\centering
\begin{adjustbox}{width=\textwidth}
\begin{tabular}{lccccccccccc}
\hline
\multirow{2}{*}{Benchmark} & \multirow{2}{*}{Size} & \multirow{2}{*}{Table} & \multirow{2}{*}{Graph} & \multirow{2}{*}{Difficulty} & \multirow{2}{*}{Answer Form} & \multicolumn{6}{c}{Problem Category} \\
\cmidrule(l){7-12}
 & & & & & & MP & CO & SO & DO & OC & GO \\
\hline
ComplexOR & 37 & {\color{red}\ding{55}} & {\color{green}\ding{51}} & {\color{red}\ding{55}} & Scalar & {\color{green}\ding{51}} & {\color{green}\ding{51}} & {\color{red}\ding{55}} & {\color{red}\ding{55}} & {\color{red}\ding{55}} & {\color{red}\ding{55}} \\
NLP4LP & 269 & {\color{red}\ding{55}} & {\color{red}\ding{55}} & {\color{red}\ding{55}} & Scalar & {\color{green}\ding{51}} & {\color{red}\ding{55}} & {\color{red}\ding{55}} & {\color{red}\ding{55}} & {\color{red}\ding{55}} & {\color{red}\ding{55}} \\
MAMO & 863 & {\color{green}\ding{51}} & {\color{green}\ding{51}} & {\color{green}\ding{51}} & Scalar & {\color{green}\ding{51}} & {\color{green}\ding{51}} & {\color{red}\ding{55}} & {\color{red}\ding{55}} & {\color{red}\ding{55}} & {\color{red}\ding{55}} \\
IndustryOR & 100 & {\color{green}\ding{51}} & {\color{red}\ding{55}} & {\color{red}\ding{55}} & Scalar & {\color{green}\ding{51}} & {\color{green}\ding{51}} & {\color{red}\ding{55}} & {\color{red}\ding{55}} & {\color{red}\ding{55}} & {\color{red}\ding{55}} \\
NL4OPT & 289 & {\color{red}\ding{55}} & {\color{red}\ding{55}} & {\color{red}\ding{55}} & Scalar & {\color{green}\ding{51}} & {\color{red}\ding{55}} & {\color{red}\ding{55}} & {\color{red}\ding{55}} & {\color{red}\ding{55}} & {\color{red}\ding{55}} \\
Optibench & 605 & {\color{green}\ding{51}} & {\color{red}\ding{55}} & {\color{red}\ding{55}} & Scalar & {\color{green}\ding{51}} & {\color{green}\ding{51}} & {\color{red}\ding{55}} & {\color{red}\ding{55}} & {\color{red}\ding{55}} & {\color{red}\ding{55}} \\
OptMATH & 166 & {\color{green}\ding{51}} & {\color{green}\ding{51}} & {\color{red}\ding{55}} & Scalar & {\color{green}\ding{51}} & {\color{green}\ding{51}} & {\color{red}\ding{55}} & {\color{red}\ding{55}} & {\color{red}\ding{55}} & {\color{red}\ding{55}} \\
CO-Bench & 6483 & {\color{red}\ding{55}} & {\color{green}\ding{51}} & {\color{red}\ding{55}} & Scalar & {\color{green}\ding{51}} & {\color{green}\ding{51}} & {\color{red}\ding{55}} & {\color{red}\ding{55}} & {\color{red}\ding{55}} & {\color{red}\ding{55}} \\
\hline
{OptiVerse-Easy} & {300} & {\color{green}\ding{51}} & {\color{green}\ding{51}} & {\color{green}\ding{51}} & {Vector} & {\color{green}\ding{51}} & {\color{green}\ding{51}} & {\color{green}\ding{51}} & {\color{green}\ding{51}} & {\color{green}\ding{51}} & {\color{green}\ding{51}} \\
{OptiVerse-Medium} & {400} & {\color{green}\ding{51}} & {\color{green}\ding{51}} & {\color{green}\ding{51}} & {Vector} & {\color{green}\ding{51}} & {\color{green}\ding{51}} & {\color{green}\ding{51}} & {\color{green}\ding{51}} & {\color{green}\ding{51}} & {\color{green}\ding{51}} \\
{OptiVerse-Hard} & {300} & {\color{green}\ding{51}} & {\color{green}\ding{51}} & {\color{green}\ding{51}} & {Vector} & {\color{green}\ding{51}} & {\color{green}\ding{51}} & {\color{green}\ding{51}} & {\color{green}\ding{51}} & {\color{green}\ding{51}} & {\color{green}\ding{51}} \\
\hline
\rowcolor{gray!20} {OptiVerse} & {1000} & {\color{green}\ding{51}} & {\color{green}\ding{51}} & {\color{green}\ding{51}} & {Vector} & {\color{green}\ding{51}} & {\color{green}\ding{51}} & {\color{green}\ding{51}} & {\color{green}\ding{51}} & {\color{green}\ding{51}} & {\color{green}\ding{51}} \\
\hline
\end{tabular}
\end{adjustbox}
\vspace{-3pt}
\caption{Comparison with existing benchmarks. OptiVerse benchmark provides structured textual contexts (table and graph), difficulty level, and vector-based solution across six optimization domains: Mathematical Programming, Combinatorial Optimization, Stochastic Optimization, Dynamic Optimization, Optimal Control, Game Optimization.}
\label{tab:benchmark_comparison}
\vspace{-2.5pt}
\end{table*}

\begin{figure*}[t]
\centering
\includegraphics[width=0.98\textwidth]{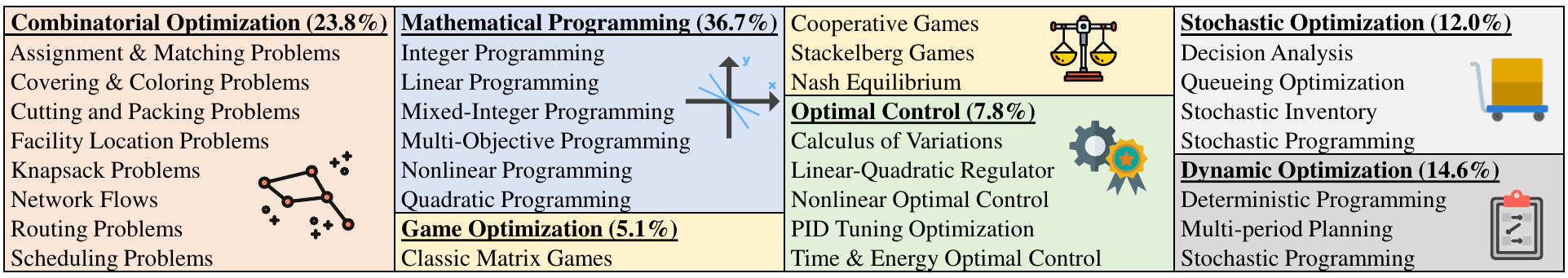}
\vspace{-5pt}
\caption{Domain distribution in OptiVerse benchmark, with each color representing a distinct optimization domain.}
\vspace{-10pt}
\label{fig:1}
\end{figure*}

\section{Related Work}
\subsection{Benchmarks for Optimization Modeling}
Evaluating LLM capabilities in optimization requires benchmarks ranging from foundational problems to complex industrial applications. Early datasets like {NL4Opt}~\cite{ramamonjison2023nl4opt} and {NLP4LP}~\cite{ahmaditeshnizioptimus} established baselines for linear and mixed-integer programming, while recent benchmarks such as {OptiBench}~\cite{yangoptibench}, {MAMO}~\cite{huang-etal-2025-llms}, and {OptMATH}~\cite{luoptmath} target advanced mathematical reasoning.  
{IndustryOR}~\cite{huang2025orlm} and {ComplexOR}~\cite{xiao2023chain} focus on practical operations research constraints. 
However, current benchmarks predominantly cover MP and CO, leaving other domains of optimization problems unexplored.

\subsection{LLM-based Optimization Modeling}
While LLMs demonstrate remarkable math reasoning capabilities, they face significant limitations in optimization problem solving involving different constraints. 
To bridge this gap, researchers develop prompt-based frameworks that utilize multi-agent workflows~\cite{xiao2023chain, ahmaditeshnizioptimus} or search algorithms~\cite{liuoptitree, astorgaautoformulation} to improve reasoning. 
Complementary to these are fine-tuning methods like FOARL~\cite{jianglarge}, ORLM~\cite{huang2025orlm}, and SIRL \cite{chen2025solver}, which train specialized models on operations research datasets to internalize complex modeling patterns.
However, prior methods often overlook modeling errors, which often manifest as silent semantic discrepancies where the logic deviates from the problem intent despite successful code execution.

\begin{figure*}[t]
\centering
\includegraphics[width=0.97\textwidth]{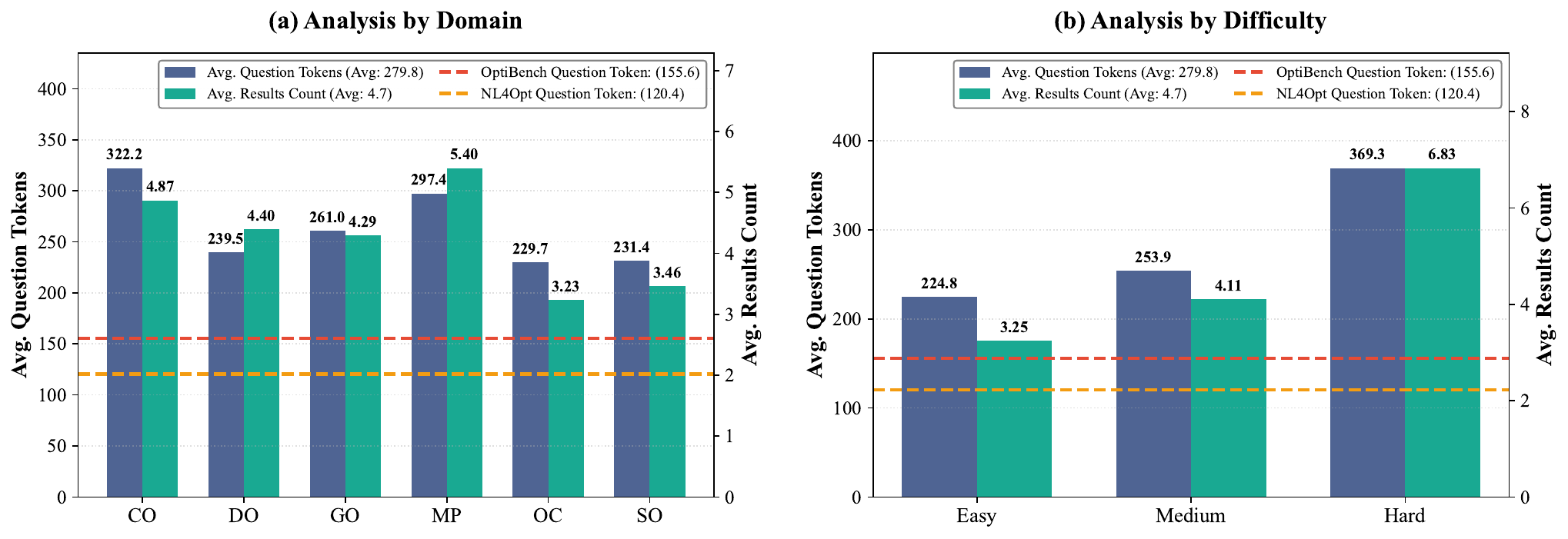}
\vspace{-5pt}
\caption{Statistical analysis of question tokens and result counts across optimization problem domain and difficulty, benchmarked against NL4Opt \cite{ramamonjison2023nl4opt} and Optibench \cite{yangoptibench}.}
\label{fig:2}
\vspace{-12pt}
\end{figure*}

\section{Benchmark}
\subsection{Data Collection and Curation}
\label{sec:3-1}
We construct OptiVerse benchmark through a comprehensive five-stage pipeline consisting of: \textbf{Acquisition}, \textbf{Standardization}, \textbf{Translation and Verification}, \textbf{Quality Filtering}, and \textbf{Classification}.
\par
\textbf{Acquisition.}
We source optimization problems from authoritative textbooks and academic publications such as~\cite{Li:2023, hillier2020intro, boyd2004convex, liu2020optimization, korte2018combinatorial, birge2011stochastic, bertsekas2017dynamic, osborne2004game} that are widely adopted in relevant curricula globally.
These span diverse optimization domains, ranging from mathematical programming to optimal control, and are selected for their rigorous academic standards and accessibility for scholarly research.

\textbf{Standardization.}
Using the MinerU2.5 framework~\cite{niu2025mineru25decoupledvisionlanguagemodel}, we extract and structure problem content from source documents. Each problem undergoes deduplication and format normalization. 
Furthermore, we ensure that tabular and graphical data are properly structured and accurately preserved alongside textual descriptions.

\textbf{Verification and Translation.}
Ph.D. candidates and master's students specializing in operations research and applied mathematics are recruited to review all instances to ensure completeness and correctness. 
Additionally, these professionals translated non-English problems to guarantee rigorous accuracy regarding domain-specific terminology, mathematical notation, and technical precision.

\textbf{Quality Filtering.} 
Following established practices~\cite{rein2024gpqa, zhang-etal-2025-physreason}, we exclude problems whose solutions are readily accessible through web searches (via a five-minute Google Search) to mitigate data contamination risks and ensure the validity and benchmark integrity.

\textbf{Classification.}
The finalized benchmark is organized into a two-dimensional taxonomy: the three aforementioned \textbf{difficulty levels} cross-referenced with six \textbf{optimization domains} as shown in Table \ref{tab:benchmark_comparison}, designed to comprehensively evaluate the ability of different models to solve optimization problems.
\par
The six domains are defined with precise boundaries: \textit{Mathematical Programming (MP)} for continuous deterministic models; \textit{Combinatorial Optimization (CO)} for discrete structures and network flows; \textit{Stochastic Optimization (SO)} for uncertainty and probabilistic models; \textit{Dynamic Optimization (DO)} for multi-stage decision processes; \textit{Optimal Control (OC)} for systems governed by differential equations; and \textit{Game Optimization (GO)} for multi-agent strategic interactions, as shown in Figure~\ref{fig:benchmark}.
This provides a comprehensive framework for evaluating the diverse reasoning capabilities of large language models, as illustrated in Figure~\ref{fig:1}.


\subsection{Benchmark Characteristics}
OptiVerse benchmark consists of 1,000 carefully curated optimization problems, providing comprehensive coverage across diverse mathematical disciplines. Figure \ref{fig:2} presents a statistical analysis of the benchmark's complexity and its distribution. 
Three distinctive features characterize this benchmark:

\textbf{Comprehensive Domain Coverage:}
Unlike existing benchmarks confined to MP and CO, OptiVerse establishes a more rigorous standard by spanning six distinct optimization domains: MP, CO, SO, DO, OC, and GO.
This coverage captures a full spectrum of challenges, ranging from static deterministic models to complex dynamic and stochastic systems. 
Consequently, this breadth ensures a holistic evaluation, testing whether LLMs possess generalized reasoning capabilities rather than merely excelling in specific sub-domains.


\textbf{Enhanced Problem Complexity and Scaling:} 
OptiVerse exhibits higher linguistic and structural complexity than existing benchmarks, with average question token and result dimensions substantially exceeding those of NL4Opt \cite{ramamonjison2023nl4opt} and OptiBench \cite{yangoptibench}. 
Furthermore, the benchmark is stratified into three difficulty levels—Easy, Medium, and Hard—where complexity scales naturally. 
For instance, Hard problems average $369.3$ tokens and require an average of $6.83$ result outputs, providing a rigorous test for complex optimization problem solving.

\textbf{Cross-Paradigm Tool Versatility:}
Given the extreme diversity of problem types—ranging from Nash Equilibrium in GO to PID tuning Optimization in OC, no single specialized solver, even a powerful industrial standard like Gurobi, can serve as a general solution for the entire benchmark.
Consequently, our dataset evaluates a model's ability to transcend single-solver dependencies. 
The LLMs must demonstrate strategic flexibility by autonomously identifying the problem's underlying paradigm and implementing the most appropriate algorithmic approach, whether through specialized libraries or custom-coded heuristic solutions.

\section{Experiments}
\subsection{Experimental Setup}
\label{sec:experimental_setup}

\paragraph{Model Selection.} 
We select 22 recent LLMs of different sizes and divide them into three categories:
\textbf{(1) Open-Source Non-Thinking Models.} 
This category encompasses the Qwen3 series (covering Instruct and Coder variants across 8B, 30B, and 235B scales) \cite{yang2025qwen3}, Qwen2.5-72B \cite{yang2024qwen2}, DeepSeek-V3.2-Chat \cite{liu2025deepseek}, Kimi-K2 \cite{team2025kimi}, Ministral3-8B \cite{mistralai2025mistral3_launch}, and Internlm3-8B \cite{internlm2025internlm3}.
\textbf{(2) Open-Source Thinking Models.} 
To evaluate reasoning capabilities within the open-weights landscape, we select OpenAI's open-sourced GPT-120B \cite{agarwal2025gpt}. Additionally, we include the specialized ``Thinking'' variants of the Qwen3 series (8B, 30B, and 235B) \cite{yang2025qwen3}, and DeepSeek-V3.2 \cite{liu2025deepseek}, which are explicitly optimized for complex reasoning tasks.
\textbf{(3) Closed-Source Thinking Models.} 
We assess proprietary systems, categorized into reasoning-intensive models—such as OpenAI's o3 and o4-mini \cite{openai2025o3o4_systemcard}, and flagship general-purpose models including GPT-5.2 \cite{openai2025gpt52}, Claude-4.5-Sonnet \cite{anthropic2025claude45}, and the Gemini series (2.5-Flash/Pro \cite{google_gemini25_flash, google_gemini25_pro} and 3-Flash/Pro) \cite{gemini3-flash,google2025gemini3}.


\paragraph{Inference Configuration.}
We adopt a chain-of-thought strategy where the model processes the problem $P$ to generate a composite response $R = \text{LLM}(P)$, consisting of a mathematical modeling component and an executable Python code segment $C$.
This formulation-first approach outperforms direct code generation by ensuring that the implementation is grounded in a formal logical derivation.
The code $C$ is then extracted from $R$ and executed within a sandboxed environment to yield the final results $O = \text{Exec}(C)$. 
To accommodate diverse optimization problem domains, our environment integrates a comprehensive suite of scientific libraries, including \textit{gurobi}, \textit{casadi}, \textit{pyomo}, \textit{nashpy}, \textit{scikit-opt}, \textit{cvxpy}, \textit{ortools}, \textit{pulp}, and \textit{scipy}.

\paragraph{Evaluation Methodology.}
To handle the complexity of diverse solver outputs and varying formatting styles, we employ a two-stage evaluation process powered by an LLM-as-judge framework.

\textbf{Stage 1: Answer Extraction.} 
In the first stage, we utilize an LLM as an extractor to parse the execution output $O$. 
The extraction process yields a corresponding set of values denoted as ${A} = \text{LLM}_{\text{extract}}(O, {R}) = \{a_1, a_2, \dots, a_n\}$, where each $a_j$ represents the extracted numerical value for the specific requirement $r_j \in R$ in problem.

\textbf{Stage 2: Answer Verification.} 
In the second stage, this judge framework performs a comparison between the extracted values $A$ and the ground truth set $A^*$. 
To account for potential minor floating-point variations inherent to solver computations, the LLM is instructed to verify precision with a relative error tolerance of $\epsilon = 0.1\%$. 
The correctness of the problem is determined by the verification of all individual components:
\begin{equation}
    \text{IsCorrect}({A}) \iff \text{LLM}_{\text{judge}}({A}, {A}^*, \epsilon) 
\end{equation}
A problem is marked as correct only when the LLM-as-judge verifies that every required variable and objective satisfies the precision threshold, thereby ensuring a rigorous assessment.

\begin{table}[t]
\centering
\begin{tabular}{lc}
\hline
\textbf{LLM} & \textbf{Accuracy (\%)} \\
\hline
Qwen3-235B-Instruct & 97.8 \\
Qwen3-235B-Thinking & 99.4 \\
DeepSeek-V3.2-Chat & 98.4 \\
DeepSeek-V3.2-Reasoner & 99.6 \\
\hline
\end{tabular}
\vspace{-3pt}
\caption{Evaluation Accuracy of Selected LLMs.}
\label{tab:llm_accuracy}
\vspace{-10pt}
\end{table}

\begin{table*}[t]
\centering
\begin{adjustbox}{width=\textwidth}
\begin{tabular}{l|cccccc|ccc|c}
\hline
\textbf{Large Language Model} & \multicolumn{6}{c|}{\textbf{Domain}} & \multicolumn{3}{c|}{\textbf{Difficulty}} & \textbf{Avg.} \\
\cline{2-10}
& MP & CO & SO & DO & OC & GO & {Easy} & {Med.} & {Hard} & \\
\hline
\rowcolor{gray!20} \multicolumn{11}{c}{\textbf{Open-Source Non-Thinking Models}} \\
Internlm3-8B-instruct & 11.44 & 7.56 & 5.00 & 8.22 & 0.00 & 3.92 & 18.67 & 3.75 & 3.00 & 8.00 \\
Ministral3-8B-Instruct & 26.01 & 18.44 & 19.35 & 12.08 & 2.56 & 5.88 & 39.33 & 14.75 & 5.67 & 18.20 \\
Qwen3-8B-Instruct & 23.98 & 22.69 & 20.83 & 14.38 & 6.41 & 13.73 & 42.67 & 12.25 & 7.67 & 20.00 \\
Qwen2.5-72B & 31.61 & 27.31 & 20.00 & 15.75 & 7.69 & 13.73 & 48.00 & 16.50 & 10.33 & 24.10 \\
Qwen3-Coder-30B & 30.79 & 31.09 & 23.33 & 23.29 & 8.97 & 9.80 & 49.67 & 19.25 & 11.67 & 26.10 \\
Qwen3-30B-Instruct & 38.96 & 36.13 & 31.67 & 34.25 & 19.23 & 15.69 & 70.33 & 21.75 & 14.00 & 34.00 \\
Kimi-K2 & 40.05 & 44.54 & 38.33 & 39.73 & 24.36 & 25.49 & 71.33 & 31.50 & 16.33 & 38.90 \\
Qwen3-235B-Instruct & 44.41 & 47.06 & 45.83 & 41.10 & 42.31 & 41.18 & 78.33 & 39.75 & 16.67 & 44.40 \\
DeepSeek-V3.2-Chat & 51.23 & 47.48 & 45.00 & 43.15 & 21.79 & 35.29 & 79.67 & 39.25 & 19.00 & 45.30 \\
\rowcolor{gray!20} \multicolumn{11}{c}{\textbf{Open-Source Thinking Models}} \\
Qwen3-8B-Thinking & 40.33 & 43.70 & 41.67 & 39.73 & 25.64 & 37.25 & 73.00 & 33.00 & 16.00 & 39.90 \\
GPT-OSS-120B & 49.54 &   55.33  &   34.19  &   38.93  &  32.05   &    35.29 & 78.67 & 39.25 & 18.67 &  44.90   \\
Qwen3-30B-Thinking & 49.59 & 46.64 & 46.67 & 46.58 & 30.77 & 43.14 & 80.67 & 40.00 & 20.33 & 46.30 \\
DeepSeek-V3.2-Reasoner & 53.68 & 52.52 & 45.00 & 49.32 & 28.21 & 49.02 & 84.33 & 44.50 & 21.33 & 49.50 \\
Qwen3-235B-Thinking & 53.68 & 51.26 & 49.17 & 52.74 & 43.59 & 49.02 & 87.33 & 47.00 & 21.33 & 51.40 \\
\hline
\rowcolor{gray!20} \multicolumn{11}{c}{\textbf{Closed-Source Thinking Models}} \\
Gemini-2.5-Flash & 46.05 & 46.22 & 48.33 & 48.63 & 48.72 & 54.90 & 82.33 & 42.75 & 18.67 & 47.40 \\
Gemini-2.5-Pro & 53.68 & 49.16 & 50.00 & 47.95 & 38.46 & 43.14 & 87.00 & 43.75 & 20.00 & 49.60 \\
Claude-4.5-Sonnet & 53.41 & 52.52 & 46.67 & 47.95 & 34.62 & 45.10 & 83.67 & 45.25 & 21.67 & 49.70 \\
o3 & 52.04 & 53.78 & 45.83 & 56.85 & 39.74 & 45.10 & 86.67 & 47.25 & 20.67 & 51.10 \\
o4-mini & 50.95 & 54.20 & 48.33 & 55.48 & 42.31 & 47.06 & 87.67 & 46.75 & 20.67 & 51.20 \\
GPT-5.2 & 55.86 & 57.14 & 50.00 & 57.53 & 50.00 & 54.90 & 91.00 & 50.75 & 25.33 & 55.20 \\
Gemini-3-Flash & 54.77 & 59.24 & 54.17 & 55.48 & 48.72 & 56.86 & 88.67 & 53.25 & 25.33 & 55.50\\
Gemini-3-Pro & 58.04 & 57.14 & 56.67 & 56.85 & 42.31 & 50.98 & 89.00 & 52.75 & 27.00 & 55.90 \\
\hline
\end{tabular}
\end{adjustbox}
\vspace{-3pt}
\caption{Model performance comparisons on the OptiVerse benchmark. The evaluation is categorized by \textbf{Domain} (MP, CO, SO, DO, OC, and GO) and \textbf{Difficulty} (Easy, Medium, and Hard). 
\textbf{Avg.} denotes the mean performance.}
\label{tab:main_results}
\vspace{-11.5pt}
\end{table*}

\begin{figure*}[t]
\centering
\includegraphics[width=0.981\textwidth]{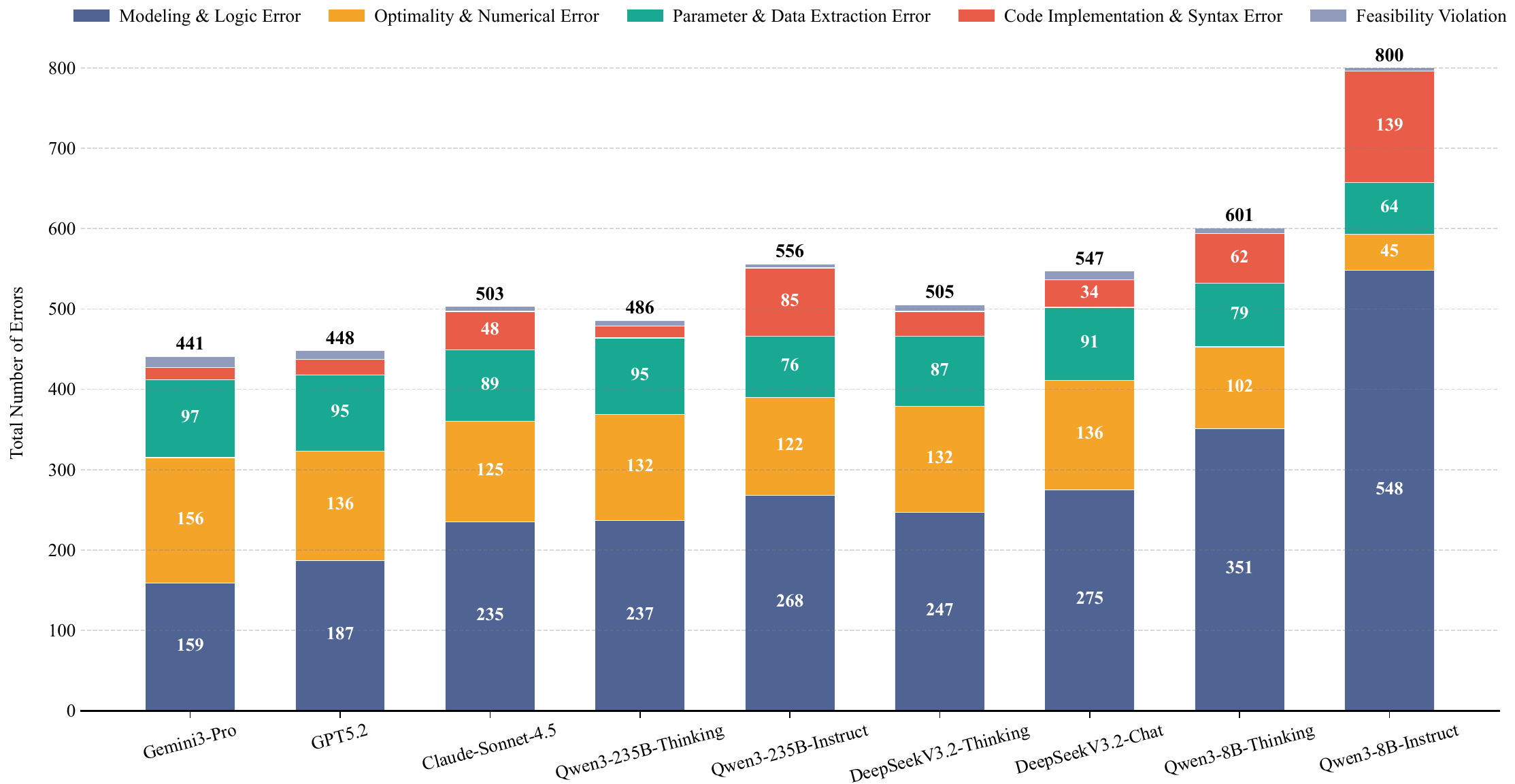}
\vspace{-1pt}
\caption{Distribution of error types across nine representative models. While \textbf{Code \& Syntax} errors become negligible with increased capability, \textbf{Modeling \& Logic} errors remain the predominant bottleneck across all LLMs.}
\label{fig:3}
\vspace{-9pt}
\end{figure*}

\paragraph{Evaluation Accuracy.}
Table \ref{tab:llm_accuracy} presents the performance statistics derived from an evaluation set of $500$ samples. 
It can be observed that advanced LLMs generally satisfy the requirements. However, considering the trade-offs between cost and inference speed, we ultimately select DeepSeek-V3.2-Chat as the standard evaluation LLM.

\subsection{Main Results}
The comprehensive evaluation results on the benchmark are presented in Table \ref{tab:main_results}. 
We summarize our key findings into four primary observations:
\par
\textbf{Superiority of Reasoning Chains:} 
Models equipped with explicit reasoning capabilities consistently outperform their standard instruction-tuned counterparts by a significant margin. 
For instance, Qwen3-8B-Thinking surpasses the performance achieved by Qwen3-8B-Instruct, which demonstrates that internal reasoning chains are crucial for addressing complex optimization problems.
\par
\textbf{Significant Difficulty Sensitivity:} 
All LLMs exhibit sharp performance degradation as task difficulty scales from Easy to Hard. 
While top-tier LLMs like Gemini-3-Pro maintain high accuracy on Easy problems, they also struggle significantly with Hard problems, indicating that LLMs struggle with solving complex optimization problems.
\par
\textbf{Scaling Law Effects:} Within the same model family, performance scales positively with parameter count. 
In the Qwen3 series, performance increases steadily from 8B to the 235B model, a trend suggesting that larger models possess more robust optimization problem-solving capabilities.

\par
\textbf{Domain-Specific Fragility:} 
There is a notable performance disparity across domains, as most LLMs perform significantly better on MP and CO tasks than on the other four categories.
This indicates that current models lack cross-domain robustness and may be sensitive to the specific symbolic representations or reasoning patterns required by different sub-fields of optimization problems.


\subsection{Error Kind Distribution Analysis}
\label{sec:error_analysis}
To identify the different failure modes impeding optimization problem solving, we conduct a fine-grained manual annotation of error types.
We categorize the failure modes into five distinct classes:
\begin{itemize}[leftmargin=*]
    \item \textbf{Modeling \& Logic Error:} The LLMs fail to correctly comprehend the problem semantics or translate the natural language description into a valid mathematical model (e.g., misinterpreting dynamic programming state transitions).
    \item \textbf{Parameter \& Data Utilization Error:} While the modeling logic is correct, the LLM fails to accurately utilize the values, coefficients, conditions, and dimensions from the text (e.g., hallucinating data, confusing data across different stages, or misaligning matrix dimensions).
    
    \item \textbf{Feasibility Violation:} The derived solution violates the problem's hard constraints, rendering the result mathematically invalid or infeasible.
    
   \item \textbf{Optimality \& Numerical Error:} The LLMs produce a feasible solution, but it is significantly suboptimal compared to the ground truth.
    
    \item \textbf{Code \& Syntax Error:} The generated code contains syntax errors, runtime exceptions, or incomplete logic, preventing successful execution
\end{itemize}

    
\begin{figure}[t]
\centering
\includegraphics[width=0.48\textwidth]{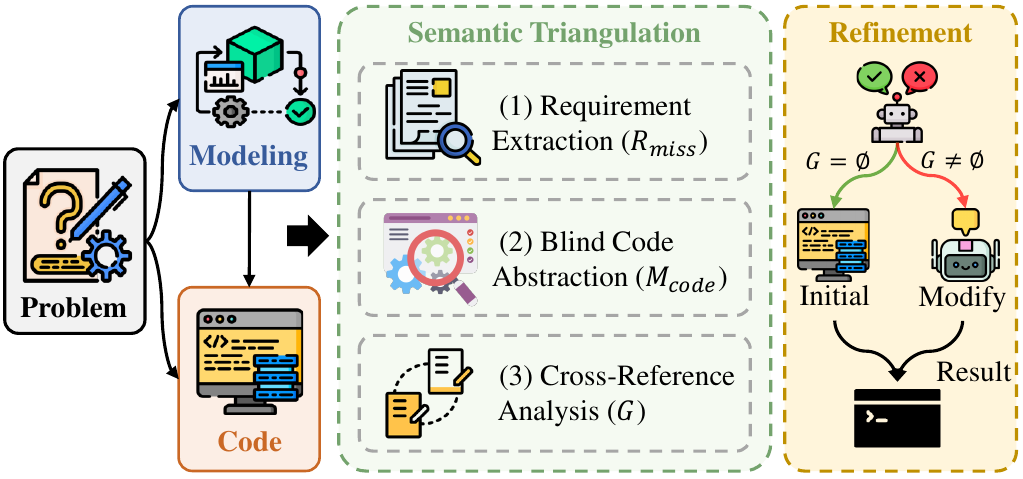}
\vspace{-12pt}
\caption{The workflow of the Dual-View Auditor Agent which utilizes a three-phase auditing mechanism to detect semantic discrepancies between the problem and code, determining whether to make modifications.}
\label{fig:agent}
\vspace{-10pt}
\end{figure}
Figure~\ref{fig:3} visualizes the comparative distribution of these error types across a diverse set of nine representative LLMs, revealing three critical insights:
\par
\textbf{Modeling is the Primary Bottleneck.} 
Even for state-of-the-art models, the prevalence of this error underscores that correctly abstracting natural language into mathematical formulations remains a persistent and fundamental cognitive bottleneck.

\textbf{Optimality \& Numerical Challenges Persist.}
After modeling errors, optimality and numerical issues constitute the second largest failure.
This suggests that even with mathematical models, LLMs often struggle to converge on optimal solutions.


\textbf{Reasoning Models Enhance Code Stability.} 
For example, the ``thinking'' variant of Qwen3-235B incurs only $15$ errors compared to $85$ in its instruct counterpart, demonstrating that extended chain-of-thought processes effectively mitigate syntax hallucinations and implementation flaws.

\section{Dual-View Auditor Agent}
\label{sec:agent}

\subsection{Method}
Motivated by the error distribution analysis in Section \ref{sec:error_analysis}, which identifies {modeling \& logic} as the predominant bottleneck in LLM-based optimization, we introduce the \textit{Dual-View Auditor Agent (DVA-Agent)}. 
Conventional self-correction methods often falter because models tend to hallucinate correctness when merely re-reading their own generated code against the prompt. 
To overcome this, DVA-Agent acts as an adversarial evaluator designed to uncover deep logical discrepancies.

\par
Central to this framework is the \textbf{Semantic Triangulation}. 
Unlike simple syntax checkers, this detects ``silent semantic errors'' scenarios where code executes without runtime exceptions but solves a mathematical model deviating from the problem intent. 
This operates through a three-phase process:

\textbf{Requirement Extraction (Text-to-Math).}
By referencing the initial modeling logic ($M_{init}$) and problem ($P$), it identifies core mathematical components, such as specific parameter values or complex constraints, that are present in $P$ but the modeling logic $M_{init}$ tends to omit or misinterpret.
This results in a set of missing requirements ($R_{miss}$) that serves as a reference for validation.

\textbf{Blind Code Abstraction (Code-to-Math).} 
To mitigate confirmation bias, the agent performs a ``blind'' review.
Without access to the problem $P$, this acts as a code interpreter, reverse-engineering the underlying mathematical logic solely from the initial code.
This extracts the objective function, decision variables, and constraints ($M_{code}$).
By isolating the code from the problem, we ensure that the derived logic reflects what the code actually executes, rather than what the LLM claims.

\textbf{Cross-Reference Analysis.} 
Finally, this synthesizes information from the three sources: the original problem $P$, the reverse-interpreted code logic $M_{code}$, and the extracted requirements $R_{miss}$.
This comparison produces a discrepancy set $G$, which quantifies the critical gap between the intended requirements and the implemented logic.

\textbf{Refinement.}
This workflow culminates in a decision-making process based on the verdict, defined by the discrepancy set $G$. 
If the analysis yields an empty set ($G = \emptyset$), signifying that the implemented code is semantically aligned with the problem, the initial code is validated and executed to compute the result. 
Conversely, if any discrepancies ($G \neq \emptyset$) are detected, the identified gaps serve as corrective guidance, prompting the LLM to modify the modeling logic and code.
This iterative refinement ensures that the final code accurately captures the complex constraints of the original problem before producing the final result. 

\begin{table}[t]
\centering
\resizebox{\columnwidth}{!}{%
\begin{tabular}{lcccc}
\toprule
Framework & {Easy}  &{Medium} & {Hard} & {Cost Time} \\
\midrule
 \rowcolor{gray!20} \multicolumn{5}{c}{\textbf{Qwen3-30B-Instruct}} \\
Baseline & {70.33}  & {21.75} & {14.00} &  6.1s\\
CoE & {70.67} & {22.50} & {14.67} & 62.4s\\
OptiMUS & {71.00}  & {24.25} & {15.33} & 35.7s\\
DVA-Agent & {72.33} & {30.50} & {18.67} & 30.2s \\
 \midrule
 \rowcolor{gray!20} \multicolumn{5}{c}{\textbf{Qwen3-235B-Instruct}} \\
Baseline & {78.33}  & {39.75} & {16.67} & 11.8s\\
CoE & {80.00} & {42.25} & {18.00} & 121.3s \\
OptiMUS & {81.33}  & {44.70} & {20.33}  & 69.3s\\
DVA-Agent & {85.67} & {50.25} & {24.33} & 58.8s\\
\midrule
 \rowcolor{gray!20} \multicolumn{5}{c}{\textbf{DeepSeek-V3.2-Chat}} \\
Baseline & {79.67} & {39.25} & {19.00} & 15.7s \\
CoE & {80.67} & {40.75} & {19.67} & 136.1s\\
OptiMUS & {81.33}  & {42.00} & {20.67} & 83.2s\\
DVA-Agent & {84.00} & {47.50} & {24.00} & 67.3s\\
\bottomrule
\end{tabular}
}
\vspace{-3pt}
\caption{Solving accuracy (\%) comparison across OptiVerse benchmark using three different LLMs. }
\label{tab:result_1}
\vspace{-10pt}
\end{table}

\subsection{Evaluation}
\label{sec:evaluation}
To rigorously evaluate the efficacy and universality of DVA-Agent, we conduct comprehensive experiments using three advanced LLMs: Qwen3-30B, Qwen3-235B \cite{yang2025qwen3}, and DeepSeek-V3.2 \cite{liu2025deepseek}.
This setup is specifically designed to validate two key advantages of DVA-Agent:
(1) \textbf{Plug-and-Play Capability}, evidenced by the seamless integration with these diverse LLMs without requiring extensive adaptation; and
(2) \textbf{Efficiency}, assessed by the impact of the conditional repair strategy that triggers regeneration only when specific discrepancies are detected.

\textbf{Baselines.}
In order to evaluate the performance against existing paradigms, we compare DVA-Agent with: Chain of Experts (CoE) \cite{xiao2023chain}, a multi-agent system utilizing domain-specific role-playing for refinement, and OptiMUS \cite{ahmaditeshnizioptimus}, a specialized framework designed for iterative optimization modeling.

\textbf{Experimental Results.} 
Table \ref{tab:result_1} empirically validates DVA-Agent's effectiveness and superiority over the existing workflow across two dimensions:
\par
\textit{(1) Robust Performance.}
DVA-Agent consistently achieves the highest accuracy across all difficulty tiers, confirming its {Plug-and-Play Capability}.
This superiority across varying model scales demonstrates that the Semantic Triangulation mechanism is model-agnostic and scales effectively with stronger reasoning backbones.
\par
\textit{(2) Computational Efficiency.}
The agent optimizes the accuracy-time trade-off via a conditional repair strategy that triggers refinement only when discrepancies arise ($G \neq \emptyset$).
With modification rates of merely 23.6\%, 32.3\%, and 28.5\% for Qwen3-30B, Qwen3-235B, and DeepSeek-V3.2-Chat, respectively, the system effectively avoids unnecessary iterations on already correct solutions.


\section{Conclusion}
We introduce \textbf{OptiVerse}, a benchmark comprising 1,000 curated optimization problems across six disciplines designed to rigorously evaluate the reasoning depth of LLMs.
Extensive experiments with 22 state-of-the-art models reveal that despite reasoning enhancements, performance on hard tasks degrades sharply, falling below 27\% accuracy.
Our fine-grained error analysis identifies semantic modeling and logic as primary bottlenecks, exacerbated by significant domain fragility when generalizing from standard mathematical programming to specialized fields like Optimal Control.
To address these deficiencies, we propose the Dual-View Auditor Agent, an adversarial framework that yields consistent improvements by targeting silent semantic defects.
Ultimately, OptiVerse provides a robust framework and actionable insights for advancing the next generation of optimization-capable LLMs.

\section{Acknowledgements}
This work was supported by Fundamental and Interdisciplinary Disciplines Breakthrough Plan of the Ministry of Education of China (JYB2025XDXM116), National Natural Science Foundation of China (No. 62137002, 62293550, 62293553, 62293554, 62437002, 62477036, 62477037, 62192781),  the Shaanxi Provincial Social Science Foundation Project (No. 2024P041), the Youth Innovation Team of Shaanxi Universities "Multi-modal Data Mining and Fusion", and Xi'an Jiaotong University City College Research Project (No. 2024Y01).

\section{Limitation}
Despite the comprehensive nature of OptiVerse, which spans six optimization domains, two key limitations warrant discussion regarding benchmark construction and evaluation.
First, our problems are primarily derived from authoritatives and academic publications, focusing on rigorous reasoning under idealized mathematical conditions rather than fully reflecting the messy realities of industrial scenarios. 
Real-world optimization often involves data noise, large-scale approximations, and ambiguous constraints that are not fully captured here. 
However, it is worth noting that mastering these rigorous formulations under idealized conditions serves as the fundamental textbook prerequisite for solving complex real-world operational problems. 
Furthermore, extensive experiments reveal that current state-of-the-art LLMs' performance on these "idealized" Hard problems remains unsatisfactory, with even GPT-5.2 and Gemini-3 failing to exceed 27\% accuracy. 
Therefore, OptiVerse remains a highly valuable testbed for evaluating models' core ability to translate natural language into executable optimization logic. 
In future research, we aim to adapt these problems to simulate real-world noise and approximation requirements. 
Second, our evaluation framework, while ensuring high precision through an LLM-as-a-judge mechanism, incurs certain computational costs. 
While the two-stage verification ensures accuracy, we plan to explore more efficient evaluation pipelines in future work to optimize the time-cost trade-off.

\section{Ethical Statement}
In developing OptiVerse, we carefully considered and addressed potential implications and risks. 
Our benchmark is sourced exclusively from public authoritative textbooks and academic publications, and we have conducted rigorous data cleansing and standardization to ensure reliability. 
To mitigate the risk of data contamination and ensure fair evaluation, we strictly excluded problems whose solutions were readily accessible through web searches. We strictly comply with the copyright and licensing terms of all source materials and the models used in our experiments. 
Committed to environmental sustainability and research reproducibility, we will publicly release the complete dataset and evaluation scripts under appropriate open-source licenses. 
This allows the community to reuse our resources, thereby cutting down the carbon footprint associated with redundant data curation, while driving the advancement of Large Language Models in complex reasoning tasks.

\bibliography{custom}

\appendix

\section{Details of Human Annotators}
For data annotation, translation, and result verification, we engaged Ph.D. candidates and Master's students who are good at Operations Research and Applied Mathematics, who are also co-authors of this paper.
All annotators possess rigorous academic backgrounds across the six optimization domains (MP, CO, SO, DO, OC, and GO), making them well-qualified to ensure the technical precision of domain-specific terminology and mathematical notations.
Since the annotators were active researchers involved in the study, no formal recruitment process or external compensation was required, and they were fully aware of the data usage protocols.
The annotation process focused solely on the evaluation of optimization problems and mathematical modeling logic, without involving the collection of any personal identifying information or exposing annotators to risks.
As this research involved the analysis of academic optimization content rather than external human subjects, it was determined to be exempt from formal ethics review board approval.
\section{Details of Ai Assistants In Research Or Writing}
We used Claude-4.5-Sonnet and Gemini-3.0-Pro to help us write code and polish the paper.

\begin{figure}[!ht]
\centering
\includegraphics[width=0.48\textwidth]{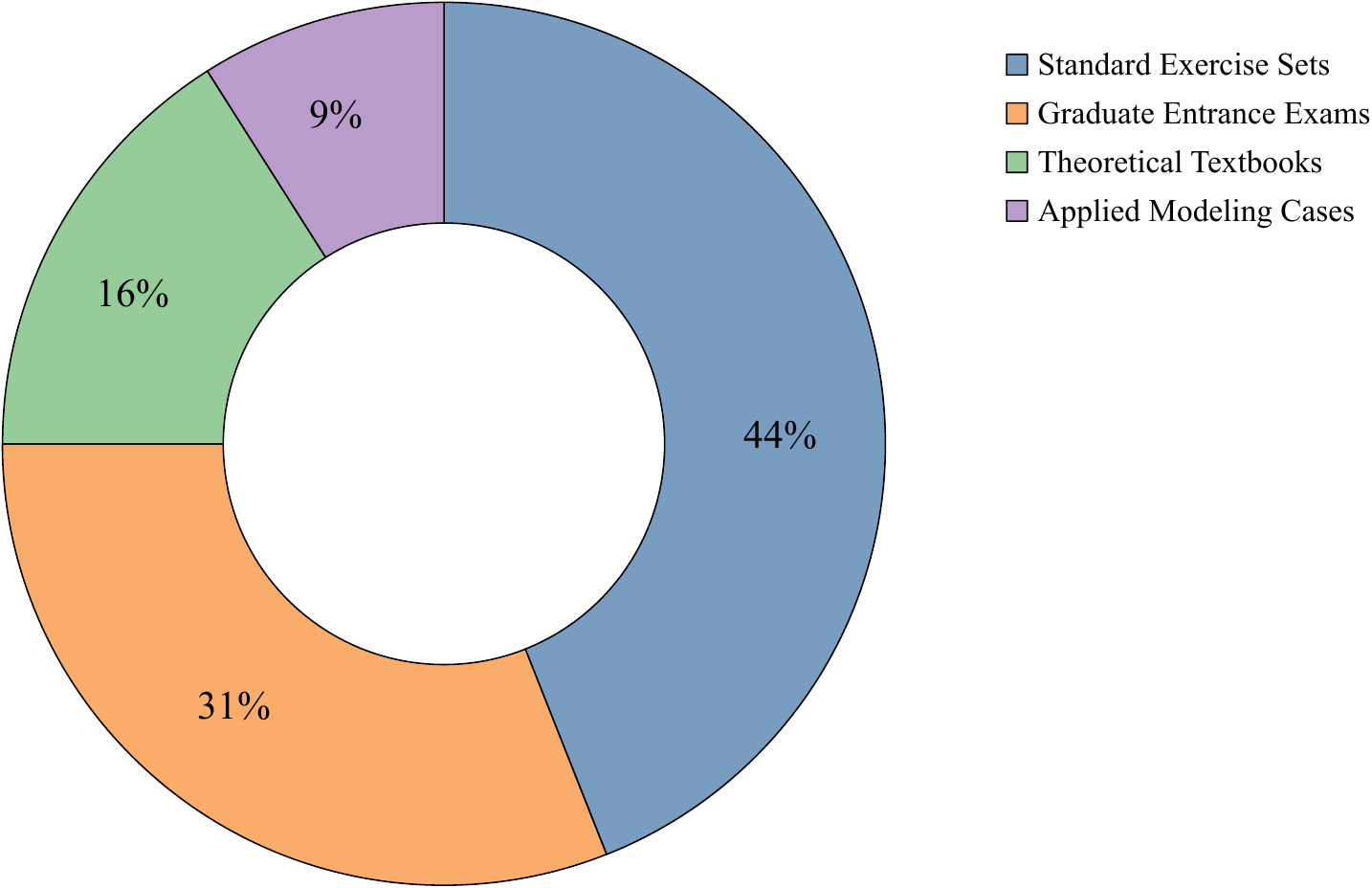}
\vspace{-13pt}
\caption{Distribution of Data Sources.}
\label{fig:Distribution of Data Sources}
\vspace{-3pt}
\end{figure}

\begin{figure}[!ht]
\centering
\includegraphics[width=0.48\textwidth]{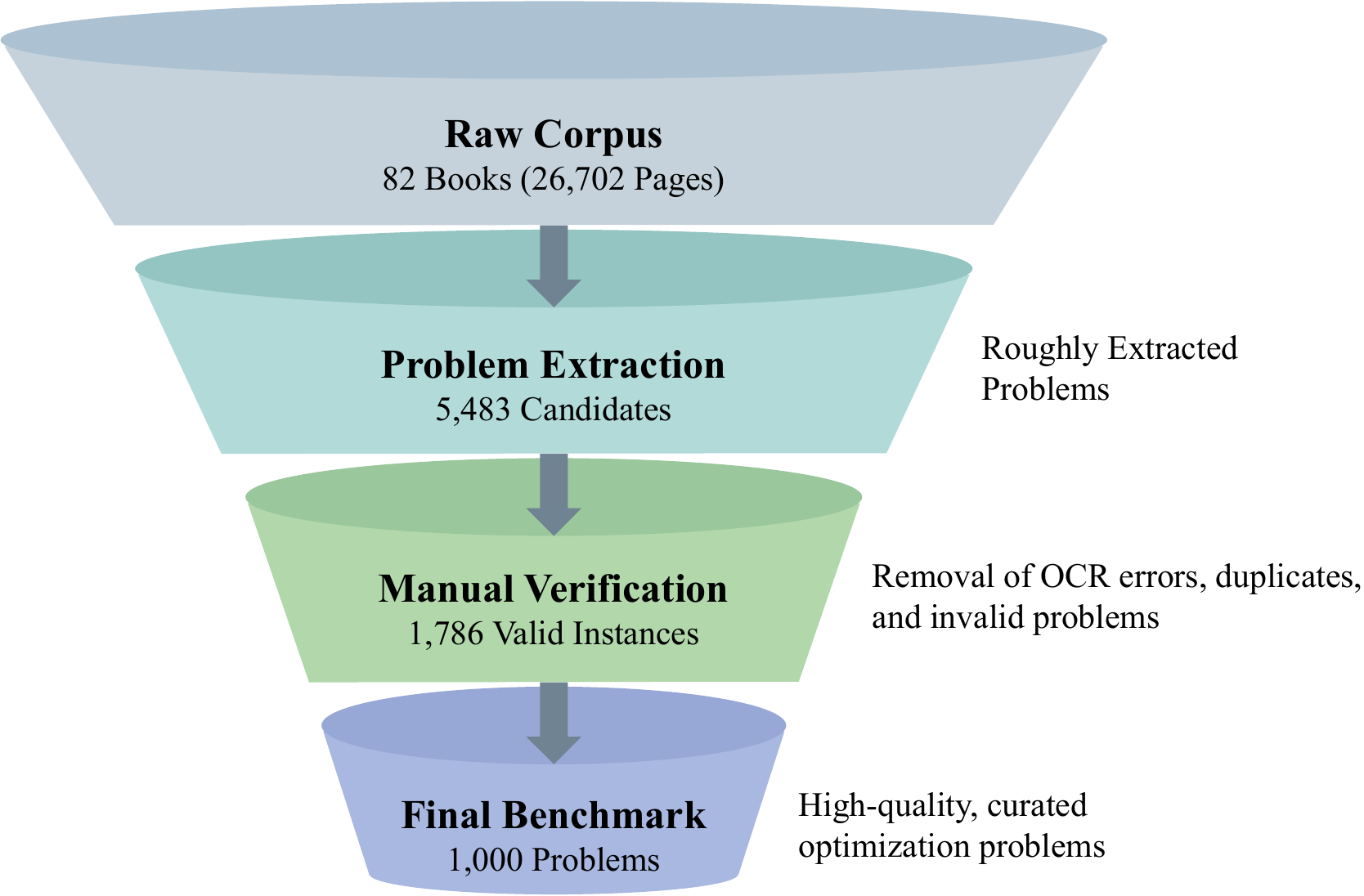}
\vspace{-12pt}
\caption{Data Construction Pipeline.}
\label{fig:Data Construction Pipeline}
\vspace{-15pt}
\end{figure}

\section{Data Source}
To construct a benchmark that is both comprehensive and rigorously challenging, we curated a massive dataset derived from highly authoritative academic resources. 
The raw corpus consists of 82 distinct textbooks, specialized problem sets, and graduate-level solution manuals, encompassing a total of 26,702 pages of source material. 
These documents cover a wide spectrum of optimization domains, ranging from fundamental Linear and Integer Programming to advanced topics in Optimal Control, Convex Optimization, and Game Theory. 
\par
Notably, to ensure the benchmark effectively evaluates the deep reasoning capabilities of Large Language Models, a significant portion of the questions was extracted from graduate entrance examination papers and mathematical modeling competition case studies, which require multi-step logical deduction rather than simple formula application.

\par
Figure~\ref{fig:Distribution of Data Sources} illustrates the distribution of these data sources by category, highlighting our focus on practical problem-solving and high-difficulty exam questions. 
\par
Figure~\ref{fig:Data Construction Pipeline} details the construction pipeline, demonstrating how we distilled the massive unstructured corpus into a refined set of 1,000 high-quality problems through a multi-stage filtration process.

\section{Error Type Details}

The detailed definitions and representative examples for each error category are presented below:
\begin{enumerate}
    \item \textbf{Modeling \& Logic Error:}
    \begin{itemize}
        \item \textit{Description:} Failures where the model correctly identifies the problem domain but fails to translate the natural language description into a valid mathematical model or optimization logic. This includes misinterpreting the objective function, omitting critical constraints, or selecting an inappropriate modeling paradigm.
        \item \textit{Examples:}
        \begin{itemize}
            \item[--] Formulating a multi-stage Stochastic Programming problem as a deterministic Linear Programming model, ignoring probability distributions.
            \item[--] In a Network Flow problem, failing to establish flow conservation constraints (i.e., inflow must equal outflow) at intermediate nodes.
            \item[--] Incorrectly defining the decision variables, such as using continuous variables for a problem that strictly requires binary or integer decisions (e.g., Knapsack Problem).
        \end{itemize}
    \end{itemize}

    \item \textbf{Parameter \& Data Extraction Error:}
    \begin{itemize}
        \item \textit{Description:} Cases where the overarching modeling logic is correct, but the model fails to accurately extract or align specific numerical values, coefficients, or dimensions from the problem text. This often manifests as "data hallucination" or matrix misalignment.
        \item \textit{Examples:}
        \begin{itemize}
            \item[--] Extracting a cost coefficient of "10" when the problem statement specifies "100", or confusing unit costs with total costs.
            \item[--] In a Traveling Salesman Problem (TSP), misreading the distance matrix indices, treating row $i$ column $j$ as the distance from $j$ to $i$ in an asymmetric graph.
            \item[--] Dimensionality mismatch, such as defining a constraint vector of size $N$ when the decision variable vector size is $N+1$.
        \end{itemize}
    \end{itemize}

    \item \textbf{Code Implementation \& Syntax Error:}
    \begin{itemize}
        \item \textit{Description:} Errors occurring during the translation of the mathematical model into executable Python code. These include syntax errors, incorrect API usage of optimization solvers (e.g., Gurobi, PuLP, SciPy), or incomplete implementation logic that causes the runtime environment to crash.
        \item \textit{Examples:}
        \begin{itemize}
            \item[--] \textbf{Syntax/Runtime Errors:} Identifying variables but failing to define them before use, or causing \verb|KeyError| or \verb|IndexError| when accessing data structures.
            \item[--] \textbf{API Misuse:} Calling non-existent methods of a solver library (e.g., using \verb|model.addConstrs| with incorrect syntax in Gurobi) or failing to call the \verb|optimize()| method.
            \item[--] \textbf{Logical Implementation Gaps:} The code runs without crashing, but the loop structure logic does not match the intended mathematical summation, resulting in an empty or trivial model.
        \end{itemize}
    \end{itemize}

    \item \textbf{Feasibility Violation:}
    \begin{itemize}
        \item \textit{Description:} This category encompasses two distinct types of failures where the output is mathematically invalid. First, it includes cases where the proposed solution explicitly violates the hard constraints defined in the problem (e.g., resource limits or non-negativity). Second, and critically, it includes cases where the generated mathematical model itself exhibits unreasonable mathematical properties—such as contradictory constraints or unbounded objectives—rendering the problem mathematically infeasible or impossible to be solved by standard algorithms.
        \item \textit{Examples:}
        \begin{itemize}
            \item[--] \textbf{Constraint Violation:} In a logistics problem, proposing a delivery route that exceeds the maximum vehicle capacity constraint, or outputting negative values for physical quantities like mass or time.
            \item[--] \textbf{Model Infeasibility:} Constructing a Linear Programming model with mutually exclusive constraints (e.g., requiring $x \ge 10$ and $x \le 5$ simultaneously), resulting in an empty feasible region.
            \item[--] \textbf{Unreasonable Mathematical Property:} Formulating an optimization objective (e.g., "Maximize Profit") without necessary bounding constraints, causing the solver to return an "Unbounded" status or diverge during calculation.
        \end{itemize}
    \end{itemize}

    \item \textbf{Optimality \& Numerical Error:}
    \begin{itemize}
        \item \textit{Description:} Cases where the model finds a valid, feasible solution, but it is significantly suboptimal compared to the ground truth, or exhibits precision issues. This often arises from inappropriate algorithm selection (e.g., using a greedy heuristic instead of an exact solver) or numerical instability.
        \item \textit{Examples:}
        \begin{itemize}
            \item[--] Using a local search algorithm (like Hill Climbing) for a non-convex function and getting stuck in a local optimum, missing the global maximum.
            \item[--] \textbf{Numerical Precision:} Returning a result that deviates from the standard solution beyond the acceptable tolerance ($\epsilon > 0.1\%$) due to floating-point errors or aggressive rounding.
            \item[--] \textbf{Solver Configuration:} Failing to set an appropriate "MIP Gap" or time limit, causing the solver to return a premature solution that has not yet converged to optimality.
        \end{itemize}
    \end{itemize}
\end{enumerate}
 
\section{Example}
\begin{figure}[t]
\centering
\includegraphics[width=0.48\textwidth]{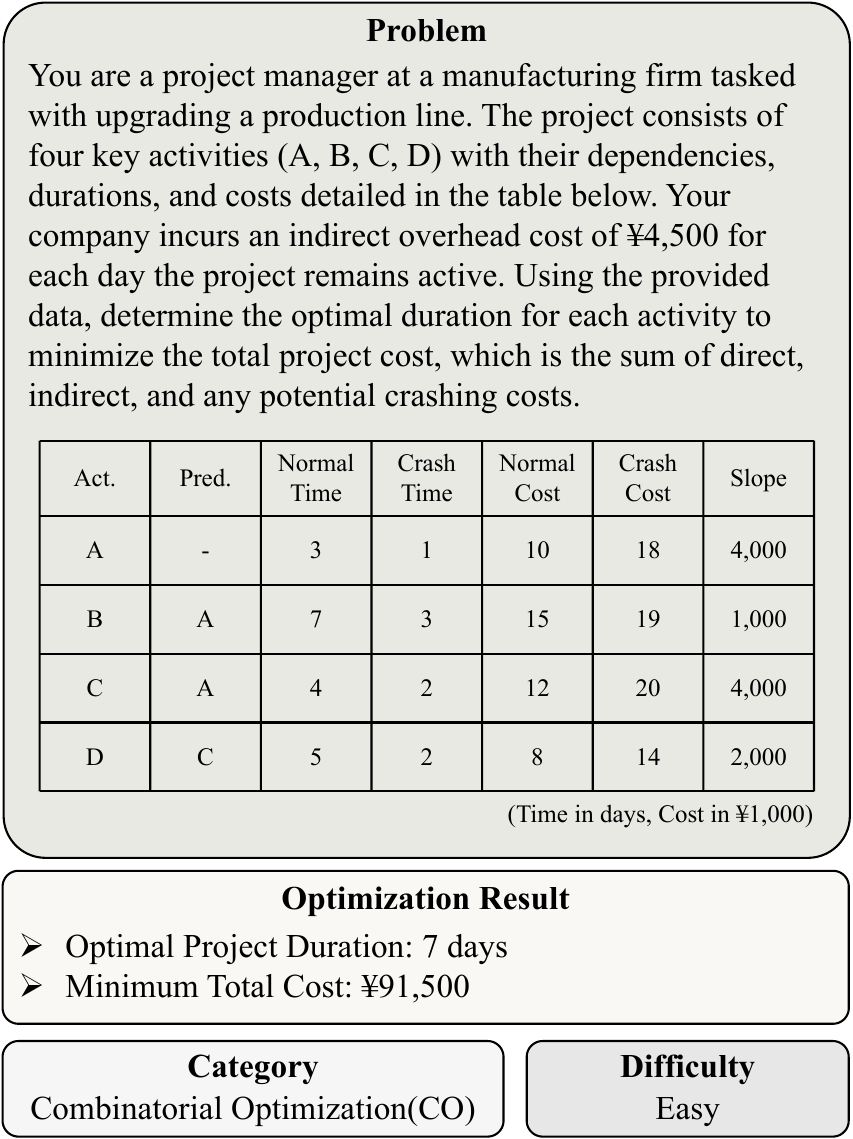}
\vspace{-10pt}
\caption{An Easy Example in OptiVerse.}
\label{fig:easy_example}
\vspace{-10pt}
\end{figure}

\begin{figure}[t]
\centering
\includegraphics[width=0.48\textwidth]{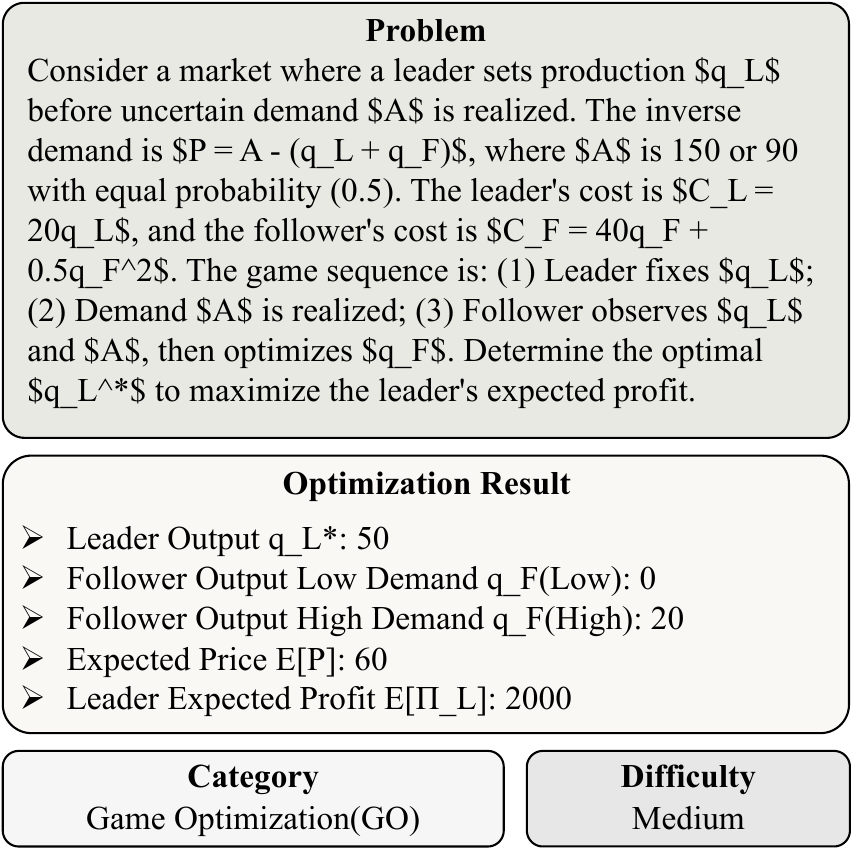}
\vspace{-10pt}
\caption{A Medium Example in OptiVerse.}
\label{fig:medium_example}
\vspace{-10pt}
\end{figure}

\begin{figure}[t]
\centering
\includegraphics[width=0.48\textwidth]{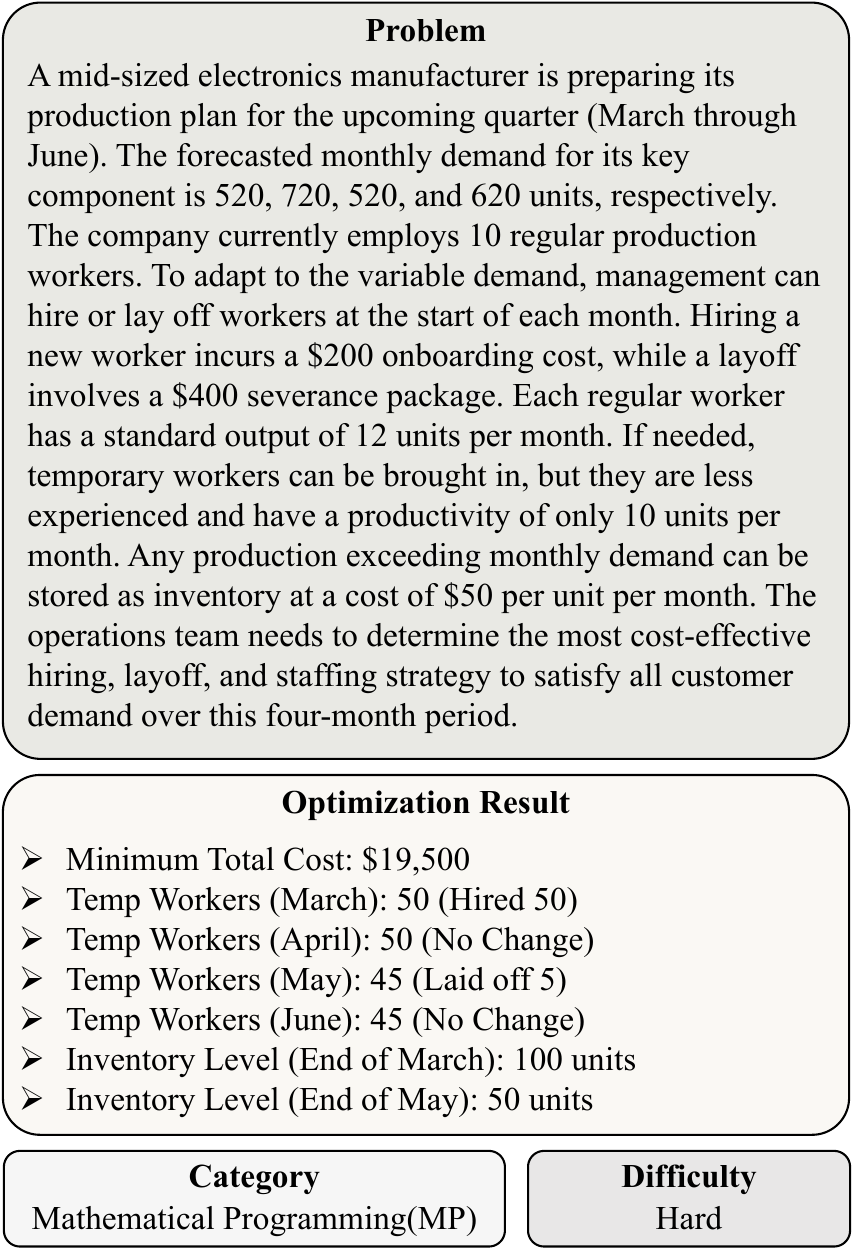}
\vspace{-10pt}
\caption{A Hard Example in OptiVerse.}
\label{fig:hard_example}
\vspace{-10pt}
\end{figure}

We have provided a representative example for each of the three difficulty levels—Easy (Figure \ref{fig:easy_example}), Medium (Figure \ref{fig:medium_example}), and Hard (Figure \ref{fig:hard_example})—to serve as a guide.
\par
The easy-level problem (Figure \ref{fig:easy_example}) represents a classic Combinatorial Optimization task involving project scheduling. It requires the model to perform algorithmic execution—specifically Critical Path Method (CPM) and cost-slope analysis. The solution path is linear and iterative: identify the critical path, select the activity with the lowest crash cost, and repeat until the optimal trade-off is found. 
This tests the model's ability to follow standard algorithmic procedures and perform precise arithmetic operations on tabular data.

The medium-level problem (Figure \ref{fig:medium_example}) introduces Game Optimization coupled with stochastic elements. 
Unlike the easy problem, it requires the integration of multiple domain concepts: Game Theory (Stackelberg Leader-Follower dynamics), Probability (Expected Value calculation), and Calculus (Profit Maximization). 
The model must employ backward induction—first solving the follower’s reaction function under different demand scenarios, and then optimizing the leader’s strategy based on expected payoffs. This demands a sophisticated understanding of multi-agent interactions and conditional logic.

The hard-level problem (Figure \ref{fig:hard_example}) presents a complex Mathematical Programming scenario involving multi-period production planning. 
The core challenge lies not just in the scale of variables (inventory, workforce, production across 4 months), but in the translation of logical business constraints into linear mathematical forms. 
Specifically, it requires modeling "hiring" and "firing" costs by splitting an unrestricted variable into two non-negative variables ($S_t = S_t^+ - S_t^-$), a modeling trick that LLMs often struggle to identify. 
The solution demands rigorous constraint handling to balance flow conservation across time periods while minimizing a composite objective function.

These examples demonstrate the progressive complexity in OptiVerse. From executing standard algorithms in Easy tasks, to synthesizing cross-domain concepts in Medium tasks, and finally to formulating complex, coupled constraints in Hard problems, the benchmark rigorously evaluates the depth of LLMs' optimization reasoning.

\section{Prompt Setting}
To ensure experimental reproducibility and the reliability of automated evaluation, we designed structured prompts that impose strict constraints on output format and reasoning logic. Unlike general-purpose queries, these prompts are engineered to bridge the gap between natural language reasoning and rigorous optimization solving.

\subsection{Inference Prompt}
The solver prompt operates under a \textbf{Code-based Chain-of-Thought} paradigm. To ensure that the generated code is executable within our sandbox environment, the prompt enforces three critical constraints:

\begin{itemize} 
\item \textbf{Expert Persona \& Tool Authorization.} The prompt establishes the model's role as an ``Operations Research Expert'' and explicitly permits access to domain-specific libraries (e.g., \texttt{scipy}, \texttt{pulp}, \texttt{ortools}, \texttt{cvxpy}). 
\item \textbf{Targeted Output Format.} To facilitate automated answer extraction, the prompt dynamically injects the required output keys (e.g., specific decision variables or objective values) and mandates that the code strictly utilizes \texttt{print()} statements to output these calculated results. 
\item \textbf{Model-then-Code Generation.} We enforce a structured reasoning process where the model is guided to first explicitly articulate the mathematical modeling logic (i.e., defining variables, constraints, and objectives) before proceeding to code implementation. Empirical results suggest that this pre-computation formalization significantly outperforms direct code generation.
\end{itemize}

\subsection{Evaluation Prompts}
We employ a two-stage prompting pipeline to ensure the robustness of the assessment, converting raw execution logs into verified verdicts.

\subsubsection{Answer Extraction Prompt}
Since the raw standard output from code execution often contains unstructured intermediate logs, we first utilize a \textbf{Structure Synthesis Prompt} to normalize the results. 
This prompt receives the raw execution output and the required schema keys from the ground truth. 
It acts as an extractor, parsing specific numerical values or decisions into a clean, valid JSON object, while handling missing values with standardized ``N/A'' markers.

\subsubsection{Accuracy Verification Prompt}
Subsequently, we employ an LLM-as-a-Judge framework to verify the correctness of the synthesized answer. 
The judge model is prompted to act as a ``Strict Mathematics Teaching Assistant,'' evaluating the ``Student's Answer'' (the synthesized JSON) against the ``Standard Solution'' based on the following criteria:

\begin{itemize} 
\item \textbf{Completeness \& Precision.} The judge verifies that all keys present in the ground truth exist in the prediction and enforces a relative numerical error tolerance of $\le 0.1\%$. 
\item \textbf{Semantic Flexibility.} The judge is instructed to intelligently handle unit variations (e.g., ``0.5'' vs. ``50\%'') and assess the semantic equivalence of non-numerical strategies (e.g., game theory decisions) rather than relying on rigid string matching. 
\item \textbf{Reasoning-First Verification.} To ensure reliability, the judge must generate a step-by-step verification log (output as the JSON field \texttt{"reason"}) prior to delivering the final boolean verdict (\texttt{"is\_correct"}).
\end{itemize}

\end{document}